\documentclass[lettersize,journal]{IEEEtran}
\usepackage[latin9]{inputenc}
\usepackage{color}
\usepackage{array}
\usepackage{multirow}
\usepackage{amsmath}
\usepackage{graphicx}

\makeatletter

\DeclareTextSymbolDefault{\textquotedbl}{T1}
\providecommand{\tabularnewline}{\\}

\let\oldforeign@language\foreign@language
\DeclareRobustCommand{\foreign@language}[1]{%
  \lowercase{\oldforeign@language{#1}}}

\usepackage{amsfonts}\usepackage{algorithmic}
\usepackage{algorithm}
\usepackage{array}
\usepackage{stfloats}
\usepackage{url}
\usepackage{verbatim}
\usepackage{cite}
\usepackage{flushend}

\makeatother

\begin{document}
\title{SADL: An Effective In-Context Learning Method for Compositional Visual
QA}
\author{
Long Hoang Dang, Thao Minh Le, Vuong Le, Tu Minh Phuong, Truyen Tran\thanks{Long Hoang Dang, Tu Minh Phuong are with the Posts and Telecommunications
Institute of Technology, Hanoi, Vietnam. Email: \{longdh, phuongtm\}@ptit.edu.vn.
Work done when Long was with Deakin.}
\thanks{Thao Minh Le, Truyen Tran are with the Applied Artificial Intelligence
Institute, Deakin University, Australia. Email: \{thao.le, truyen.tran\}@deakin.edu.au.}
\thanks{Vuong Le is with the Amazon, Melbourne, Australia. Work done when
Vuong was with Deakin. Email: levuong@amazon.com.}
}

\maketitle



\global\long\def\ModelName{\textrm{SADL}}%
\global\long\def\LVLMs{\textrm{LVLMs}}%

\begin{abstract}
Large vision-language models (LVLMs) offer a novel capability for
performing in-context learning (ICL) in Visual QA. When prompted with
a few demonstrations of image-question-answer triplets, LVLMs have
demonstrated the ability to discern underlying patterns and transfer
this latent knowledge to answer new questions about unseen images
without the need for expensive supervised fine-tuning. However, designing
effective vision-language prompts, especially for compositional questions,
remains poorly understood. Adapting language-only ICL techniques may
not necessarily work because we need to bridge the visual-linguistic
semantic gap: Symbolic concepts must be grounded in visual content,
which does not share the syntactic linguistic structures. This paper
introduces $\ModelName$, a new visual-linguistic prompting framework
for the task. $\ModelName$ revolves around three key components:
\textbf{SA}mpling, \textbf{D}eliberation, and Pseudo-\textbf{L}abeling
of image-question pairs. Given an image-question query, we sample
image-question pairs from the training data that are in semantic proximity
to the query. To address the compositional nature of questions, the
deliberation step decomposes complex questions into a sequence of
subquestions. Finally, the sequence is progressively annotated one
subquestion at a time to generate a sequence of \emph{pseudo-labels}.
We investigate the behaviors of $\ModelName$ under OpenFlamingo on
large-scale Visual QA datasets, namely GQA, GQA-OOD, CLEVR, and CRIC.
The evaluation demonstrates the critical roles of sampling in the
neighborhood of the image, the decomposition of complex questions,
and the accurate pairing of the subquestions and labels. These findings
do not always align with those found in language-only ICL, suggesting
fresh insights in vision-language settings. 
 
\end{abstract}


\section{Introduction}

Large vision-language models (LVLMs) have recently achieved state-of-the-art
in in-context learning (ICL) across various vision-language tasks
\cite{alayrac2022flamingo,bubeck2023sparks,Tsimpoukelli2021MultimodalFL,Chen_2023_ICCV,liu2023llava}.
In visual question answering (VQA), this is achieved by prompting
the LVLMs with a few image-question-answer triplets. This success
is particularly important as it is data efficient and allows for straightforward
inference-only deployment without the need for costly tuning \cite{10295530}.
Nevertheless, there is still a lack of understanding in effectively
devising vision-language prompts, especially for complex compositional
questions.

In language-only settings, textual prompts have shown effectiveness
when carefully sampled examples are used \cite{levy2023diverse,liu-etal-2022-makes,rubin2022learning},
along with deliberation steps like those found in Chain-of-Thought
(CoT) \cite{wei2022chain} and Least-to-Most (L2M) \cite{zhou2023leasttomost},
which break down the reasoning process into a chain of steps. However,
it remains uncertain whether these techniques will translate to vision-language
settings, where prompts involve both visual content and interleaved
text. Visual QA, in particular, requires grounding linguistic concepts
in visual content, extracting relevant information related to each
step of reasoning \cite{hudson2018compositional,le2020dynamic,johnson2017clevr,9762027,10171793,10550013,10483073,9826805}.
These steps cannot be easily described in linguistic form as in language-only
settings. In our preliminary evaluations of these techniques, CoT
and L2M offered only minimal differences compared to vanilla few-shot
methods as in \cite{alayrac2022flamingo,li2023blip,li2023mimic} on
large-scale compositional VQA datasets. This result suggests the necessity
for new multimodal prompting approaches.

\begin{figure}
\begin{centering}
\includegraphics[width=0.9\columnwidth]{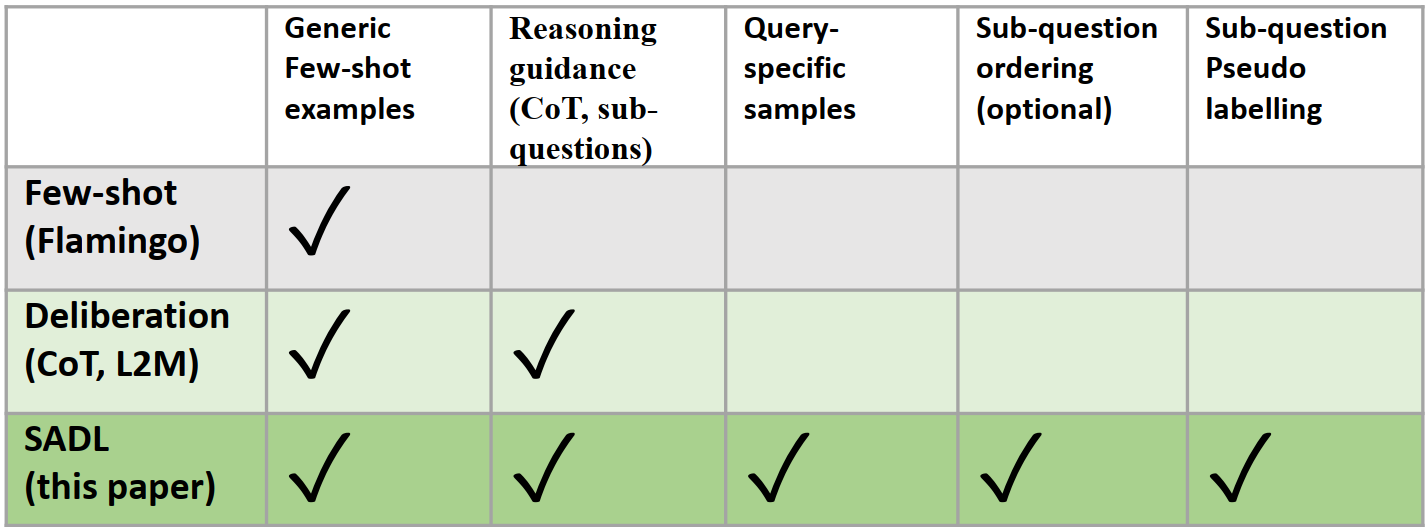}
\par\end{centering}
\caption{We study the prompting techniques for vision-language reasoning tasks
to bridge the gap of compositional reasoning and LVLM inference-free
reasoning.\label{fig:what-we-offer}}
\end{figure}

Directly aiming at this challenge, this paper embarks a study on the
scheme of\emph{ }ICL using\emph{ guided visual-linguistic prompting
techniques. }We design\emph{ a prompting framework }called\emph{ $\ModelName$
to guide the inference of LVLMs in compositional Visual QA tasks }and\emph{
examine the effectiveness of this inference-only scheme.} This framework
revolves around three key components: \textbf{SA}mpling, \textbf{D}eliberation,
and structured Pseudo-\textbf{L}abeling of image-question pairs. When
faced with an image-question query, $\ModelName$ first samples a
few semantically related image-question pairs from the training data.
This is hypothesized to enable LVLMs to recognize the focused reasoning
patterns by learning \emph{in-context} from demonstrations of similar
visual content and questions. To handle the compositional nature of
questions, we implement the deliberation step, inspired by recent
works in LLMs \cite{khot2023decomposed,zhou2023leasttomost}, which
prompts an LLM to break down complex questions into a sequence of
subquestions. Subsequently, the sequence is successively annotated
by the same LVLM, one subquestion at a time, resulting in the generation
of a sequence of \emph{pseudo-labels}. See Fig.~\ref{fig:what-we-offer}
to see how $\ModelName$ contributes to the state of the art.

We evaluate $\ModelName$ using the OpenFlamingo framework \cite{anas_awadalla_2023_7733589}
on large-scale Visual QA datasets, namely GQA \cite{hudson2019gqa},
GQA-OOD \cite{kervadec2021roses}, CLEVR \cite{johnson2017clevr},
and CRIC \cite{9905976}. The results demonstrate three significant
advantages of this guided prompting scheme:

1. Inference-only without fine-tuning, hence readily to use with out-of-the-box
LVLMs, alleviating the heavy expertise and work load required in supervised
training.

2. Maintaining a reliable performance on visual-linguistic compositional
reasoning tasks, improving over state-of-the-art few-shot methods
such as CoT and L2M.

3. Demonstrating a harmonic middle-ground hybrid of supervised-learning
and few-shot learning where annotated labels and pseudo-labels are
used effectively in a training-free inference-only scheme.

Key factors for success include sampling within the neighborhood of
the image-question query, decomposing complex questions, and accurately
pairing subquestions and pseudo-labels. These observations diverge
from those found in language-only ICL, providing valuable insights
in the context of vision-language settings.
{\small{} }{\small\par}

\section{Related work}

\paragraph{Inference of LVLMs in VQA}

Recent studies have shown that the large vision-language models (LVLMs)
exhibit the in-context learning capability in the Visual Question
Answering (VQA) task, yielding promising results. Tsimpoukelli et
al. (2021) introduced a novel approach to training a vision encoder
for a static language model, resulting in a multimodal system that
can extend its proficiency to new vision-language tasks \cite{Tsimpoukelli2021MultimodalFL}.
Alayrac et al. (2022) combined the potential of extensively pretrained
vision and language models to rapidly adapt to the VQA task, using
only a limited set of examples \cite{alayrac2022flamingo}. Yang et
al. (2022) encoded the image into captions or tags, incorporating
them into the few-shot examples, then asked GPT-3 to answer questions
about the given images \cite{yang2022empirical}. More recently, the
latest large-scale multimodal model, GPT-4, as evidenced in \cite{openai2023gpt4,bubeck2023sparks}
has showcased the capability to process both image and text as inputs,
displaying strong performance under an inference-only setting. However,
there remains a big gap in the literature concerning effective methods
for prompting LVLMs, especially for compositional VQA. We address
this challenge by focusing on the behaviors of our proposed $\ModelName$
framework under OpenFlamingo \cite{anas_awadalla_2023_7733589} using
extensive large-scale Visual QA datasets to extract insights in vision-language
settings.

\paragraph{Demonstrations with deliberation steps }

These prompting methods have proved beneficial. Chain-of-thought
prompting \cite{wei2022chain} learns the reasoning process by breaking
down the mapping from input to output into a chain of steps. To avoid
the manual annotating limitation, Zhang et al. (2022) leveraged a
prompt ``\emph{Let's think step by step}'' to automate the generation
of chain-of-thought structures \cite{zhang2022automatic}. Along the
same lines, Press et al. (2022) asked LLMs to deconstruct the complex
question into simpler ones, addressing these sub-questions before
combine them with a search engine to create the prompt \cite{press2022measuring}.
Wang et al. (2022) proposed an innovative approach, iCAP, that can
adjust the context of the prompt in every deliberation step \cite{wang2022iteratively}.
Additionally, Zhou et al. (2023) introduce a prompt strategy that
comprising two stages: query LLMs to decompose complex problem and
solving these subproblems sequentially \cite{zhou2023leasttomost}.
$\ModelName$ pushes along this line by decomposing compositional
questions into a succession of subquestions, and ensuring precise
alignment between subquestions and corresponding labels.

\paragraph{Demonstrations selection}

The shared objective among all prompting strategies falling under
the demonstration selection category is to retrieve ``high-quality''
examples for the ICL process. Liu et al. (2022) utilize nearest neighbor
algorithm to select the ICL examples that highly relevant to the text
query \cite{liu-etal-2022-makes}. Beside distance metrics, mutual
information has been used to seek the ``good'' prompt from a set
of candidate prompt \cite{sorensen-etal-2022-information}. Similarly,
the selection of top-$k$ cross-lingual demonstrations has been employed
to enhance the alignment between source and target inputs \cite{tanwar-etal-2023-multilingual}.
Additionally, Gonen et al. (2022) \cite{gonen2022demystifying} shown
that choosing low perplexity prompts can boost the performance of
LLMs model. Levy et al. (2023) focused on retrieving demonstrations
that contribute to the diversity of ICL examples \cite{levy2023diverse}.
While these efforts have yielded promising results, they are limitted
in the language-only setting. In constrast, our proposed $\ModelName$
selects query-specific demonstrations that target visual-linguistic
queries whose language-visual content relations play a crucial role.

\section{Method}

\begin{figure*}
\begin{centering}
\includegraphics[width=0.9\textwidth]{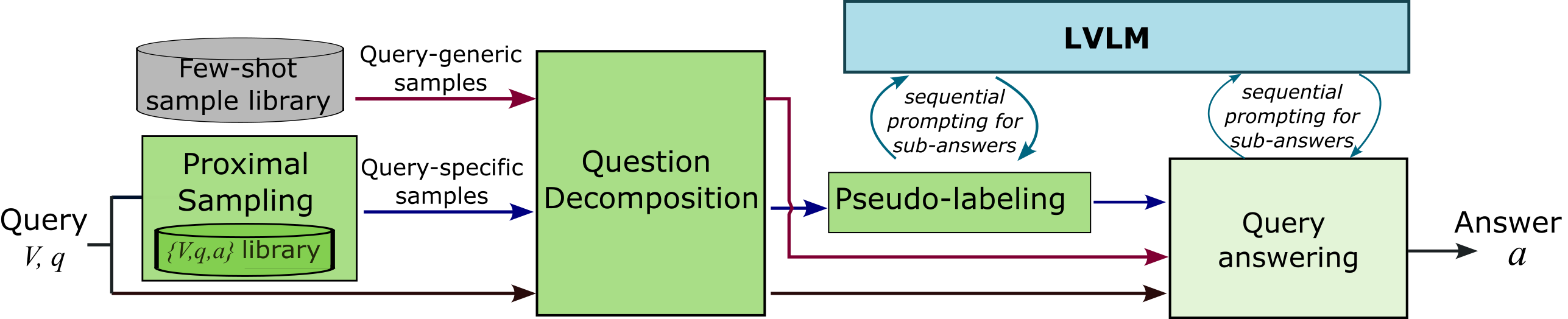}
\par\end{centering}
\caption{Overview of the $\protect\ModelName$ framework. Advancing the direction
of in-context learning for VQA, we introduce three new components
(dark green boxes) for effective compositional reasoning using LVLM.
The \emph{Proximal Sampling} uses L-V similarity to select the \emph{query-specific
samples. }Their prioritized list of sub-questions are constructed
by \emph{Question Decomposition} and used to guide the compositional
reasoning in the \emph{Pseudo-labeling} step, providing the effective
context for query answering. \label{fig:content}}
\end{figure*}

\subsection{Preliminaries}

We study a VQA setting in which a pre-trained Large Vision Language
Model (LVLM) is parameterized by $\theta$. This model is prompted
by a question $q$ regarding a visual content $v$ given a context
$C$. The LVLM then returns an answer $a^{*}$ sampled from the conditional
distribution:
\begin{equation}
a^{*}=\arg\max_{a\in A}P_{\theta}\left(a\mid C,v,q\right)\label{eq:VQA-LVLM}
\end{equation}
In this paper, we will focus on in-context learning (ICL) scenarios,
where $C$ consists of $k$ demonstrations in the form:

\begin{equation}
C=\left\{ c_{i}\mid c_{i}\equiv\left(v_{i},q_{i},r_{i},a_{i}\right);\text{for}\,i=1,2,...,k\right\} \label{eq:ICL-context}
\end{equation}
where $r_{i}$ is optional deliberation step. Here the model parameter
$\theta$ remains unchanged, but the model behaviors are shaped by
the demonstrations, implicitly modeling the underlying patterns in
the context \cite{mirchandani2023large}. For instance, in the context
of \emph{chain-of-thought} \cite{wei2022chain}, $r_{i}$ is a sequence
of reasoning steps to arrive at $a_{i}$.

It has been shown recently that the sampling and formatting of $C=\{c_{i}\}$
play a critical role in deriving the performance of the LLMs \cite{lu-etal-2022-fantastically,honovich2022instruction,wu2023selfadaptive,tanwar-etal-2023-multilingual}.
However, little is known about how to design an effective context
$C$ in vision-language systems.  A straightforward idea is to borrow
techniques confirmed in LLMs into this new multimodal setting. To
test the idea, we perform a preliminary study on how recent language
prompting techniques such as vanilla few-shot, Chain-of-Thought (CoT)
\cite{wei2022chain} and Least-to-Most (L2M) \cite{zhou2023leasttomost}
work on VQA. Particularly, we conducted exploratory experiments using
an open-source LVLM called OpenFlamingo \cite{anas_awadalla_2023_7733589}
in a two-shots setting on the three large-scale compositional VQA
datasets: GQA \cite{hudson2019gqa}, CLEVR \cite{johnson2017clevr},
and CRIC \cite{9905976} (details in the Experiments section). The
results (See Table \ref{tab:Performance-on-GQA}, \ref{tab:Performance-on-CLEVR},
and \ref{tab:Performance-on-CRIC}) show that more deliberate prompts
such as CoT and L2M only slightly improve over the vanilla few-shot
by less than 3 percent-points on all datasets, and L2M is not better
than CoT. This result contradicts with the findings on language-only
similar tasks. \emph{The discrepancy suggests that linguistic prompting
techniques that work on language-only setting do not necessarily translate
to vision-language settings.} We conjecture that in VQA, encoded visual
content is not readily mapped into discrete concepts to allow straightforward
visual grounding via standard self-attention mechanisms found in Transformer-based
LVLM architectures. Thus fresh multimodal prompting methods are required.

\subsection{$\protect\ModelName$}

We now present an effective in-context learning method with a novel
strategy for \textbf{\emph{SA}}\emph{mpling, }\textbf{\emph{D}}\emph{eliberation,
and structured Pseudo-}\textbf{\emph{L}}\emph{abeling }(abreviated
as \emph{$\ModelName$})\emph{ }to guide the LVLM for this VQA task.
The focus of the method is on the coupling between the visual content
and compositional linguistic questions, implemented as sampling strategies
to construct the context $C=\left\{ c_{i}\right\} _{i=1}^{k}$ in
Eq.~(\ref{eq:ICL-context}). Drawing from the previous generally
transferable insights in LLMs, we hypothesize that the visual-linguistic
context $C$ is effective when it contains demonstrations that are
both (a) \emph{task-representative}--those demonstrating exemplar
reasoning patterns in solving the task, but not necessarily semantically
proximal to the query and the content, and (b) \emph{query-specific}--those
semantically similar to the current query and content. Intuitively,
the former may allow the LVLM to draw from the syntactic analogy between
semantically distal instances. On the other hand, the latter makes
the context $C$ semantically coherent, thereby sharpening the distribution
of internal ``topics'' recognized by the LVLM \cite{xie2021explanation},
which helps generate better answers.

Yet another crucial aspect is the compositionality of the question,
which can be dealt with by breaking it into smaller subquestions which
are easier to answer. We argue that question decomposition is even
more important in visual-linguistic settings because it will be much
easier for the LVLM to reason about a single aspect of the visual
content, which is itself typically compositional (e.g., a scene is
typically composed of objects and their relations, and an object is
composed of parts).

Toward this goal, we design $\ModelName$ with three new steps: (1)
Sample selection, (2) Question decomposition, and (3) Subquestion
(pseudo-)labeling. Fig.~\ref{fig:content} provides an overview of
the $\ModelName$ prompting techniques. 

\subsection{Generic-specific sample selection}

For Step 1, the $\ModelName$ employs two sampling strategies for
demonstrations -- generic and specific with respect to the visual
content and the question. 

\textbf{Query-generic samples}: Representative demonstrations highlight
inference patterns that one may use to answer a question about a randomly
chosen image, which are not necessarily related to the target image.
In practice, the patterns can be in the form of chain-of-thoughts
\cite{wei2022chain} or a set of sub-problems generated from the pre-defined
complex problem \cite{zhou2023leasttomost,khot2023decomposed}. We
use a small set of $k_{1}$ such patterns, shared among all test queries.

\textbf{Query-specific samples}: For each image/question pair, query-specific
demonstrations are drawn from semantically similar visual content
and questions. In particular, given an image-question pair $(v,q)$
from a testing set, we select a small set of $k_{2}=k-k_{1}$ most
similar samples from a library of samples. In practice, we use training
sets of respective datasets as the library. It is worth mentioning
that while the\textbf{ }image often contains a significant amount
of redundant information, the question does not. Therefore, we first
search for samples in the training set with the most similar question
to the testing question $q$. Among the samples with similar questions,
we then pick the ones whose associated images are most similar to
the testing image $v$. To represent the semantics of images, we use
image captions generated by a LVLM model \cite{anas_awadalla_2023_7733589}
or high-level features extracted by Vision Transformer \cite{Dosovitskiy2020AnII}.
The degree of similarity is measured by cosine distance between the
questions and images in feature space. 

\subsection{Question decomposition}

This step translates each sample question $q_{i}$ into a set of $m$
subquestions $\left\{ \hat{q}_{i,1},\hat{q}_{i,2},...,\hat{q}_{i,m}\right\} $.
This is done by prompting an external LLM using few-shot learning,
where each shot has the format: \emph{``To answer the question:}
{[}compositional question{]}, \emph{we need to know}: {[}subquestion
1{]}, {[}subquestion 2{]}, ... {[}subquestion m{]}''.

\begin{figure}[ph]
\includegraphics[width=1\columnwidth]{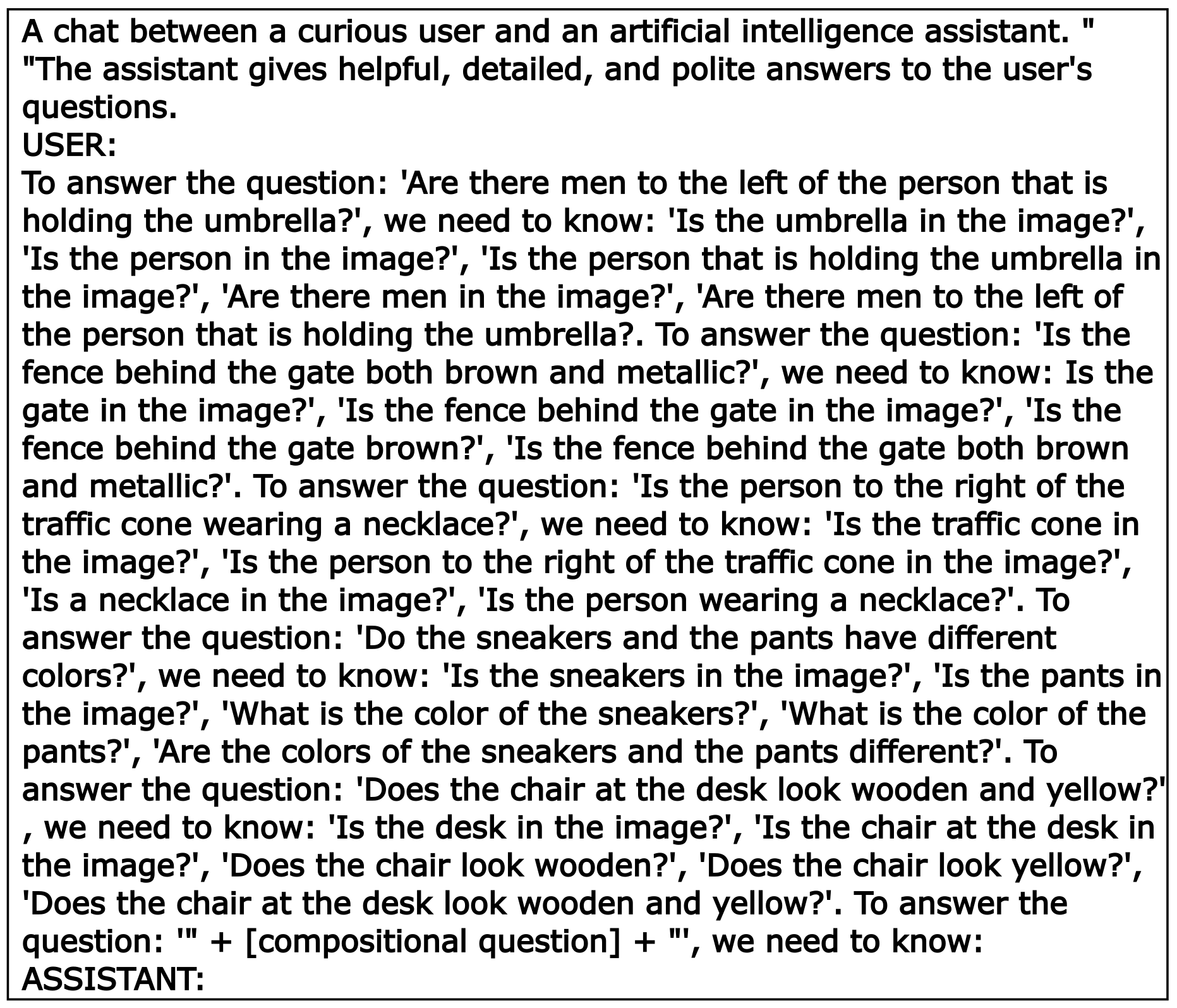}

\caption{Example of our prompt to instruct the Vicuna 13B model to decompose
the complex compositional question. \label{fig:Examples}}
\end{figure}

As an illustration of the question decomposing process, consider the
following question: \emph{``Are there men to the left of the person
holding the umbrella?}'', we prompt the Vicuna 13B model \cite{vicuna2023}
to return a set of subquestions: \emph{``Is the umbrella present
in the image?}''; \emph{``Is there a person present in the image?}'';
\emph{``Is the person holding the umbrella?}''; \emph{``Are there
men present in the image?}''; and \emph{``Are there men positioned
to the left of the person holding the umbrella?}''. Fig. \ref{fig:Examples}
displays our prompt for the question decomposition step.

Inspired by the intuition that human often find it more productive
to answer easy questions first and gain insight to answer more difficult
questions \cite{dua2022successive}, we can optionally orders the
resulting subquestions (and corresponding answers when suitable) by
their increasing difficulty level. To estimate the difficulty level
of each subquestion, we employ the Berkeley Neural Parser \cite{Kitaev2018MultilingualCP}
to parse each subquestion and obtain its syntactic tree. The level
of difficulty to each subquestion is then assigned based on the number
of significant noun phrases in its parsed tree. For example, the question
\textquotedbl\emph{Are there}\textbf{\emph{ men}}\emph{ positioned
to the left of}\textbf{\emph{ the person}}\emph{ holding }\textbf{\emph{the
umbrella}}\emph{?}\textquotedbl{} has the level of difficulty of 3
because it contains three significant noun phrases as as indicated
in bold text.

\subsection{Subquestion (pseudo-)labeling}

Once a sample question is decomposed into a series of subquestions,
we annotate each subquestion with an answer. For query-generic samples,
the answers are manually given. On the other hand, for query-specific
samples, the answers are auto-generated, i.e., pseudo-labeling by
applying the\emph{ }adjustment procedure. The automated annotation
proceeds as follows. Given an initial prompt including the set of
query-generic samples, and sample image $v_{i}$, the LVLM is asked
to generated an answer $\hat{a}_{i,1}$ for the simplest subquestion
$\hat{q}_{i,1}$. Then, this subquestion $\hat{q}_{i,1}$ and its
generated answer $\hat{a}_{i,1}$ are appended to the prompt, and
the LVLM will predict the answer for the second subquestion $\hat{q}_{i,2}$.
The iterative process continues until the answer $\hat{a}_{i,m}^{\textrm{ }}$
to the last subquestion $\hat{q}_{i,m}$ is obtained. If $\hat{q}_{i,m}$
is answered correctly ($\hat{a}_{i,m}^{\textrm{ }}=$ $a_{i,m}$),
the model proceeds to the next stage. If not, the ground truth answer
$a_{i,m}$ of the question $\hat{q}_{i,m}$ is provided to the LVLM,
and it is compelled to answer the subquestions from the beginning.
This iterative adjustment procedure ensures that the LVLM refines
its predictions for pivotal subquestions, mimicking a step-by-step
reasoning process to arrive at the correct answer.

\subsection{Query answering}

Finally, we ask the LVLM to answer the testing question $q$ about
the testing image $v$. Here, the question $q$ will be decomposed
into a set of subquestions. Then, the LVLM is called to answer the
chain of testing subquestions sequentially, utilizing both query-generic
and query-specific samples. As the LVLM's output is unconstrained,
we need to semantically align the output with one of given possible
answers in the answer set $A$ of Eq.~(\ref{eq:VQA-LVLM}). More
specifically, we align to the closest word in $A$ using cosine similarity.

\section{Experiments}

We implement our proposed prompting technique $\ModelName$ with the
pretrained OpenFlamingo\footnote{https://github.com/mlfoundations/open\_flamingo}
\cite{anas_awadalla_2023_7733589} as the LVLM. We follow the same
setting as in prior studies \cite{zhou2023leasttomost}, where the
LVLM are provided with $k=2$ demonstrations. 

\subsection{Datasets}

We evaluate the effectiveness of $\ModelName$ through comprehensive
experiments across various compositional VQA benchmarks. These benchmarks
include GQA \cite{hudson2019gqa}, GQA-OOD \cite{kervadec2021roses},
CLEVR \cite{johnson2017clevr}, and CRIC \cite{9905976}. We intentionally
choose these benchmarks as they are created to justify different aspects
of machine learning models and often necessitate the use of design-specific
supervised methods \cite{hudson2018compositional,hudson2019learning,Nguyen2021CoarsetoFineRF}:

\textbf{GQA} comprises over 22M compositional questions, derived from
over 110K images from Visual Genome \cite{Krishna2016VisualGC}. It
is currently the most extensive benchmark for compositional visual
reasoning with a diverse set of questions. These questions are challenging
as they require different reasoning skills such as spatial reasoning,
relational reasoning, logic and comparison. 

\textbf{GQA-OOD }is a out-of-distribution variant derived from the
GQA dataset. It consists of nearly 54K questions associated with 10K
images. Answering questions in this dataset not only requires the
same reasoning capability as it does in the GQA data but also the
ability to overcome the distribution shifts between train and test
splits. 

\textbf{CRIC dataset }offers 494K balanced and compositional questions,
accompanied by 96K real images from Visual Genome. Unlike the questions
in GQA and GQA-OOD, the questions in this dataset focus more on the
ability to understand visually grounded commonsense knowledge.

\textbf{CLEVR }includes 860K questions annotated from 100K synthetic
images of 3D objects. Thanks to the synthetic nature of the data,
it serves as a true out-of-distribution setting to test the generalization
capability of LVLMs as they are never exposed to data of this type.
Additionally, questions in this dataset are more complex as they contain
higher order of object's relationships.

\subsection{Computing infrastructure}

$\ModelName$ is implemented using Python 3.9.16 and Pytorch 2.0.0.
All experiments in this paper are conducted utilizing GPU NVIDIA A100
SXM4 40GB with total 320 GB of memory, running on Ubuntu 20.04.4 LTS.

\subsection{Baseline prompting techniques}

\begin{figure}
\includegraphics[width=1\columnwidth]{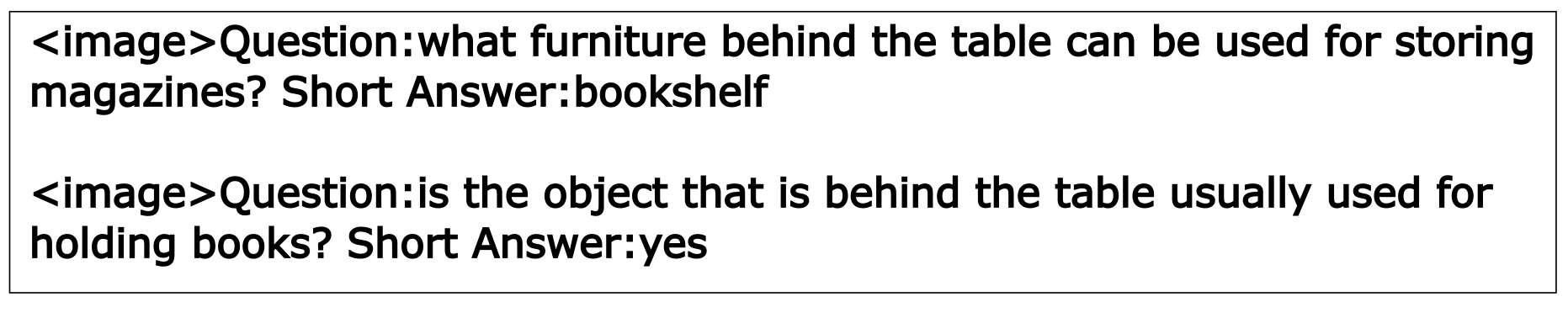}

\caption{Example of vanilla few-shot prompting (two-shot setting) \label{fig:Example-of-vanilla}}
\end{figure}

\textbf{Vanilla few-shot prompting. }For the first baseline, we examine
the vanilla few-shot prompting \cite{alayrac2022flamingo}, wherein
the model seeks to identify an answer $a^{*}$ for a question $q$
that is grounded with an image $v$, given in-context samples $C$
structured as image-query-answer triplets $C=\left\{ c_{i}\mid c_{i}\equiv\left(v_{i},q_{i},a_{i}\right);\text{for}\,i=1,2,...,k\right\} $.
Fig. \ref{fig:Example-of-vanilla} displays an example prompt.

\begin{figure}
\includegraphics[width=1\columnwidth]{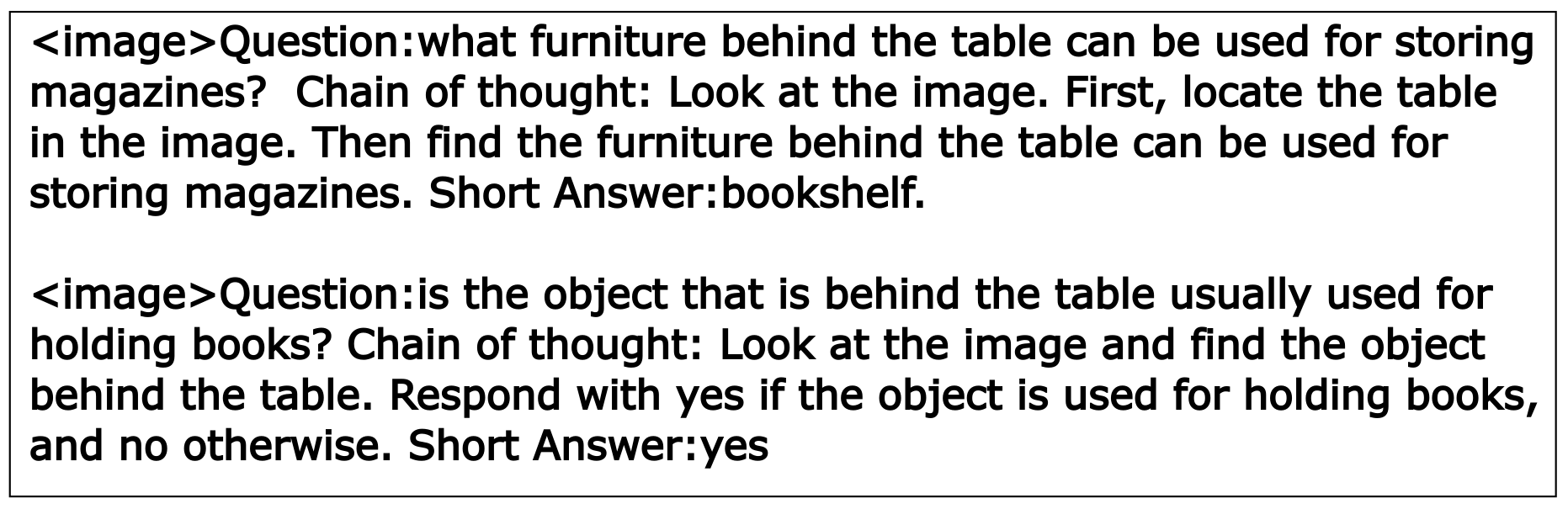}

\caption{Example of chain-of-thought prompting (two-shot setting) \label{fig:Example-of-CoT}}
\end{figure}

\textbf{Chain-of-thought. }For the second baseline, we examine the
chain-of-thought prompting \cite{wei2022chain}, wherein the in-context
samples $C$ is represented as:

\begin{equation}
C=\left\{ c_{i}\mid c_{i}\equiv\left(v_{i},q_{i},r_{i},a_{i}\right);\text{for}\,i=1,2,...,k\right\} \label{eq:ICL}
\end{equation}
where $r_{i}$ is a sequence of reasoning steps to arrive at $a_{i}$.
Fig. \ref{fig:Example-of-CoT} displays an example prompt.

\begin{figure}
\includegraphics[width=1\columnwidth]{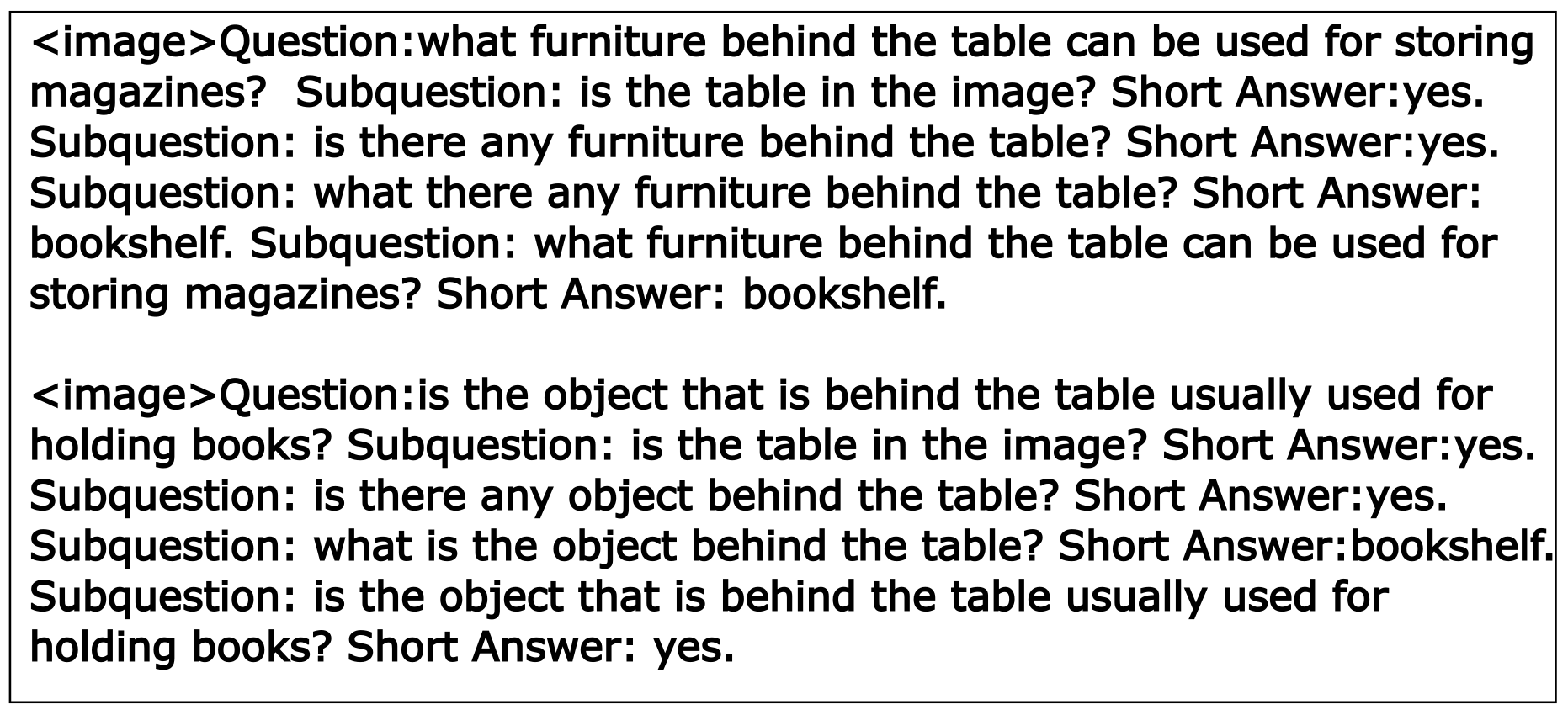}

\caption{Example of least-to-most prompting (two-shot setting) \label{fig:Example-of-L2M}}
\end{figure}

\textbf{Least-to-most }For the third baseline, we examine the least-to-most
prompting \cite{zhou2023leasttomost}, where $r_{i}$ in Eq. \ref{eq:ICL}
is a set of sub-problems generated from the pre-defined complex problem.
Fig. \ref{fig:Example-of-L2M} displays an example prompt.

\subsection{Comparison against SOTAs \label{subsec:Comparison}}

\begin{table*}
\centering{}\caption{Experimental results on CRIC. (Obj = Object; KG = Knowledge Graph;
SG = Scene Graph; Att = Attribute) \label{tab:Performance-on-CRIC}}
\begin{tabular}{l|c|c|c|c|c|c}
\hline 
\multirow{2}{*}{{\small{}Model}} & \multicolumn{2}{c|}{KG-specific questions} & \multicolumn{3}{c|}{Scene-specific questions} & \multirow{2}{*}{Overall$\uparrow$}\tabularnewline
\cline{2-6} \cline{3-6} \cline{4-6} \cline{5-6} \cline{6-6} 
 & QueryObjKG$\uparrow$ & VerifyKG$\uparrow$ & QueryObjSG$\uparrow$ & QueryAtt$\uparrow$ & VerifyAtt$\uparrow$ & \tabularnewline
\hline 
{\small{}Vanilla prompting} & 27.65 & 37.07 & 16.82 & 18.16 & 22.99 & 24.55\tabularnewline
{\small{}Chain-of-thought (CoT)} & 24.02 & 39.49 & 13.56 & 21.10 & 36.64 & 25.19\tabularnewline
{\small{}Least-to-Most (L2M)} & 27.08 & 32.32 & 18.89 & 24.18 & 28.21 & 25.69\tabularnewline
\textbf{\small{}$\ModelName$ (Ours)} & \textbf{42.85} & \textbf{55.54} & \textbf{27.30} & \textbf{30.38} & \textbf{47.00} & \textbf{39.47}\tabularnewline
\hline 
\end{tabular}
\end{table*}

\begin{table*}[t]
\centering{}\caption{Experimental results on GQA. \label{tab:Performance-on-GQA}}
\begin{tabular}{l|c|c|c|c|c|c|c}
\hline 
\multirow{1}{*}{{\small{}Model}} & Binary$\uparrow$ & Open$\uparrow$ & Consistency$\uparrow$ & Validity$\uparrow$ & Plausibility$\uparrow$ & Distribution$\downarrow$ & Accuracy$\uparrow$\tabularnewline
\hline 
{\small{}Vanilla prompting} & 41.64 & 19.85 & 9.60 & 67.55 & 63.56 & 30.79 & 30.39\tabularnewline
{\small{}Chain-of-thought (CoT)} & 42.93 & 22.18 & \textbf{12.16} & 71.18 & 67.07 & 31.32 & 32.22\tabularnewline
{\small{}Least-to-Most (L2M)} & 46.96 & 18.78 & 10.21 & 69.48 & 65.77 & 35.09 & 32.42\tabularnewline
\textbf{\small{}$\ModelName$ (Ours)} & \textbf{56.10} & \textbf{26.19} & 8.97 & \textbf{92.47} & \textbf{87.66} & \textbf{7.68} & \textbf{40.67}\tabularnewline
\hline 
\end{tabular}
\end{table*}

\begin{table}[t]
\centering{}\caption{Experimental results on GQA-OOD; ``acc'' is short for accuracy.
\label{tab:Performance-on-GQA-OOD}}
\begin{tabular}{l|c|c|c}
\hline 
\multirow{1}{*}{{\small{}Model}} & acc-all{\footnotesize{}$\uparrow$} & acc-tail{\footnotesize{}$\uparrow$} & acc-head{\footnotesize{}$\uparrow$}\tabularnewline
\hline 
{\small{}Vanilla prompting} & 25.39 & 21.15 & 27.53\tabularnewline
{\small{}Chain-of-thought (CoT)} & 27.69 & 22.65 & 30.25\tabularnewline
{\small{}Least-to-Most (L2M)} & 26.60 & 20.96 & 29.45\tabularnewline
\textbf{\small{}$\ModelName$ (Ours)} & \textbf{35.46} & \textbf{27.82} & \textbf{39.33}\tabularnewline
\hline 
\end{tabular}
\end{table}

\begin{table}[t]
\centering{}\caption{Experimental results on CLEVR. \label{tab:Performance-on-CLEVR}}
\begin{tabular}{l|c}
\hline 
\multirow{1}{*}{{\small{}Model}} & Accuracy$\uparrow$\tabularnewline
\hline 
{\small{}Vanilla prompting} & 34.38\tabularnewline
{\small{}Chain-of-thought (CoT)} & 37.04\tabularnewline
{\small{}Least-to-Most (L2M)} & 37.15\tabularnewline
\textbf{\small{}$\ModelName$ (Ours)} & \textbf{41.14}\tabularnewline
\hline 
\end{tabular}
\end{table}

\begin{table}[t]
\centering{}{\footnotesize{}\caption{Model behaviors against design choices on CRIC. \label{tab:Design-choices}}
}%
\begin{tabular}{l|c}
\hline 
{\footnotesize{}Model} & {\footnotesize{}Val. Acc.$\uparrow$}\tabularnewline
\hline 
{\footnotesize{}Full model} & {\footnotesize{}39.47}\tabularnewline
\hline 
{\footnotesize{}- random, non-specific demonstrations} & {\footnotesize{}27.81}\tabularnewline
{\footnotesize{}- w/o question decomposition} & {\footnotesize{}26.54}\tabularnewline
{\footnotesize{}- randomizing position of pseudo-labels} & {\footnotesize{}30.46}\tabularnewline
{\footnotesize{}- fixed pseudo-labels as ``unknown''} & {\footnotesize{}16.83}\tabularnewline
\hline 
\end{tabular}
\end{table}

We compare $\ModelName$ against recent language prompting techniques
such as vanilla few-shot prompting, Chain-of-Thought prompting (CoT)
\cite{wei2022chain}, and Least-to-Most (L2M) prompting \cite{zhou2023leasttomost}
on the mentioned datasets. More details about these techniques are
available in the supplementary material.  

\textbf{Results on CRIC:} Table~\ref{tab:Performance-on-CRIC} presents
experimental results on commonsense questions on CRIC dataset. It
is clear that $\ModelName$ significantly outperforms the others by
large margins on all question types. Among these question types, we
observe that knowledge graph related questions are beneficial the
most from our sample selection, as the knowledge retrieved from our
provided samples are more transferable to answer these questions.
These other question types, including ``QueryObjSG'', ``QueryAtt''
and ``VerifyAtt'', are more scene-specific questions hence, the
improvements are mostly from better visual grounding capabilities
offered by $\ModelName$. We provide more insights on the effects
of each designed component of $\ModelName$ on the overall performance
later in the ablation studies. 

\textbf{Results on GQA:} Table~\ref{tab:Performance-on-GQA} details
the experiment results on the GQA validation set. As shown, $\ModelName$
surpasses all the considered baselines across various metrics by large
margins. Particularly, $\ModelName$ achieves an overall accuracy
of 40.67\%, gaining an improvement of over 8.0 absolute points compared
to CoT and L2M. This improvement is evenly distributed across both
binary and open-ended questions. More importantly, we observe unprecedented
improvements regarding ``validity'', ``plausibility'' and ``distribution''
that are on par with supervised methods \cite{hudson2019gqa}. We
believe this is a direct result of our local sample selection that
facilitates LVLM to draw topic-focused analogies relevant to the question
scope (See our Discussion section). It is to note that like all other
prompting methods, ours also suffers from hallucinations and inconsistent
responses, resulting in poor ``consistency'' of only around 10.0\%,
while SoTA supervised methods reportedly score over 80.0\% on this
metric \cite{hudson2019gqa}. This suggests a need for future studies
in this space.

\textbf{Results on GQA-OOD:} Interestingly, $\ModelName$ achieves
even more significant improvement over the other prompting methods
in out-of-distribution settings, as evidenced in Table~\ref{tab:Performance-on-GQA-OOD}.
Specifically, it gains around 28.0\% performance increase compared
to CoT overall with a 30.0\% relative improvement of prediction of
the head distribution of answers and 22.8\% relative improvement of
the tail distribution of answers. We speculate this is a benefit of
our subquestion pseudo-labeling that provides better local reasoning
patterns to infer rare concepts (See more in our Discussion section).

\textbf{Results on CLEVR:} Our proposed $\ModelName$ also outperforms
all other prompting methods when answering questions in the CLEVR
dataset, although with smaller gap than the other real-world datasets.
Experimental results on this dataset is provided in Table~\ref{tab:Performance-on-CLEVR}.
Particularly, $\ModelName$ achieves 41.14\% overall accuracy on validation
set, nearly 4.0 points gained from L2M. These results once again confirm
the effectiveness of our prompting technique in solving compositional
questions, even on synthetic images.

\begin{figure}
\begin{centering}
\includegraphics[width=0.9\columnwidth,height=0.8\columnwidth]{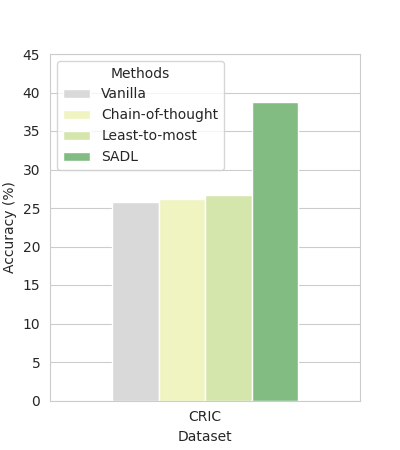}
\par\end{centering}
\caption{Performance of $\protect\ModelName$ against the other few-shot prompting
techniques on a set of multi-step compositional questions in CRIC.
\label{fig:plot}}
\end{figure}

\begin{figure*}
\centering{}\includegraphics[width=0.8\textwidth]{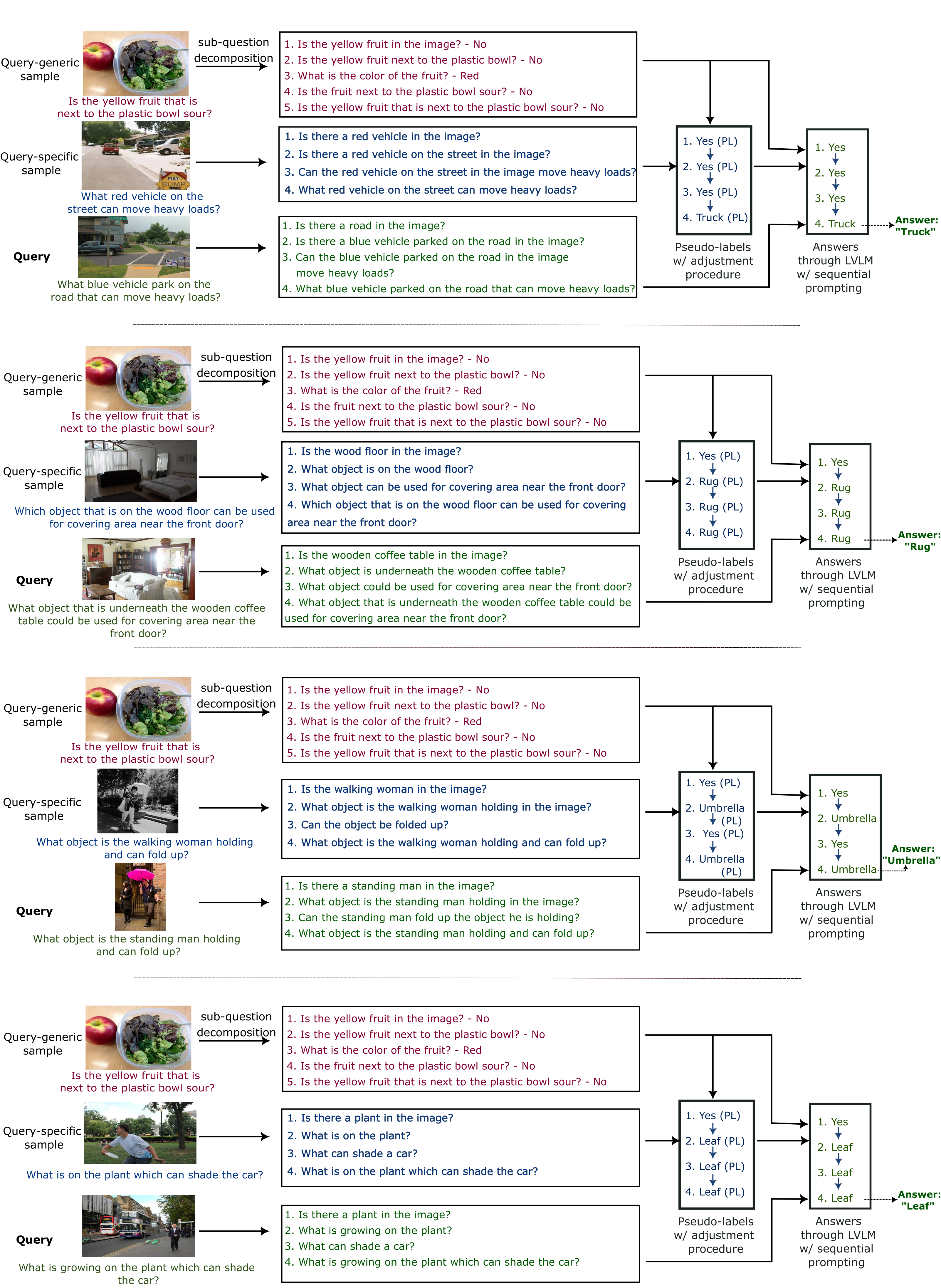}\caption{Illustration of the dataflow inside $\protect\ModelName$ prompting
framework on an example query. Query-specific and query-generic sample
questions are decomposed into respective sets of subquestions. They
guide the process of autonomous pseudo-labels generation and adjustment,
leading LVLM to answer the query correctly. \label{fig:Qualitative-analysis}}
\end{figure*}

\subsection{What make $\protect\ModelName$ work?}

To gain deeper insights into the three-step design of $\ModelName$,
we conduct ablation studies on design choices using the CRIC dataset
(See Table~\ref{tab:Design-choices}). Overall, skipping any of
the proposed steps would negatively impact the overall performance.
Our ablation experiments include:

\textbf{\emph{Random, non-specific demonstrations:}}\textbf{ }We replace
the query-specific samples by a random subset of samples drawn from
the training set to observe the impact of the query semantic proximity.
This leads to a significant decrease in performance, by nearly 12.0
points, equivalently a 29.5\% relative drop. The result confirms the
role of using similar in-context samples in the few-shot prompting
setting.

\textbf{\emph{W/o question decomposition:}} Without decomposing questions,
the performance is drastically down to 26.54\% from 39.47\% of the
full method. The result aligns with prior studies \cite{press2022measuring}
and again confirms the critical role of this deliberation step in
the ICL demonstration towards answering compositional questions.

\textbf{\emph{Randomizing position of pseudo-labels}}\textbf{: }To
test whether exact pairing of sub-questions and pseudo-labels is crucial,
we shuffle the positions of auto-generated pseudo-labels. This results
in a significant drop of 9 points, or a 22.8\% decrease from the full
method. The result reveals that observations in language-only ICL
are not necessarily generalizable to vision-language large models
as \cite{min2022rethinking} has found that exact input-output pairing
is not required (Also see the third point in our Discussion section).

\textbf{\emph{Fixed pseudo-labels}}\textbf{ }\textbf{\emph{as ``unknown''}}\textbf{:
}To further validate the impact of quality pseudo-labels of subquestions
on the performance, we conduct another ablation study where prefix
them as all ``unknown''. These noisy labels seem to spoil the contextual
information to $\LVLMs$ and mislead the predictions, resulting in
a poor performance of only 16.83\%. This again confirms the strong
correlation between the quality of pseudo-labels and correct responses
by $\LVLMs$.

\subsection{Performance on highly complex questions}

To assess the advantages of our proposed $\ModelName$ in answering
complex compositional questions, we conduct an experiment using a
subset of highly intricate questions extracted from the CRIC dataset.
These questions necessitate a minimum of five reasoning steps to reach
the answers. The findings, as shown in Fig.~\ref{fig:plot}, clearly
demonstrate that $\ModelName$ outperforms current SoTA methods by
large margins, i.e., 45\% relative gain over L2M, when tackling complex
questions. These results strongly evidenced the efficacy of $\ModelName$
in facilitating multi-step compositional reasoning tasks in comparison
with existing approaches.

\subsection{Visualization}

Fig.~\ref{fig:Qualitative-analysis} provides a representative example
that illustrates the decomposed demonstrations produced through our
proposed $\ModelName$ prompting framework. It clearly showcases how
$\ModelName$ leverages relevant information from query-specific samples,
decomposes complex questions, iteratively generates high-quality pseudo-labels,
and gathers all this contextual information to arrive at the correct
answer.

\section{Conclusion}

In-context learning (ICL) is a recent phenomenon, thought to be an
``emergent ability'' found in sufficiently large language models
\cite{wei2022emergent} and vision-language models \cite{alayrac2022flamingo}.
Consequently, we have not yet established a widely accepted understanding
of the underlying principles driving the effectiveness of ICL in language-only
settings; and our understanding in vision-language contexts is very
limited. Our work contributes in this direction by investigating design
choices for visual-linguistic prompting, tested on compositional Visual
QA. The findings include: 

1. \textbf{Local sample selection:} Random sampling of demonstrations
was found to hurt performance substantially. While this outcome aligns
with earlier observations in language-only scenarios, we anticipate
the impact to be even more pronounced in Visual QA due to the distinctly
content-centric nature of the task. Compositional questions necessitate
a sophisticated deliberation by decomposing the answering process
into multiple reasoning steps, each of which asks for language grounding
and language-induced visual relations \cite{hudson2018compositional,le2020dynamic}.
When related samples are introduced in the prompt, their deliberations
might facilitate LVLMs to draw topics-focused analogies about these
step-by-step language-vision interactions.

2. \textbf{Deliberation:} Question decomposition by prompting an LLM
was a key to $\ModelName$'s performance, although we did not perform
extensive search on the best prompting format. This is inline with
the ``self-ask'' technique used by \cite{press2022measuring} but
differs in the way we used a separate LLM for the step. \textcolor{black}{We
note that our question decomposition is a problem-solving strategy,
but it does not result in a precise chain of logical reasoning. While
answering easier questions might give some early clues and insights,
compositional question might benefit more from parallel collecting
of evidences -- the topics/Bayesian assumption \cite{xie2021explanation}.
Also, the underlying neural architecture of LVLMs, the Transformer,
does not depend on orders.}

3. \textbf{The structured pseudo-label space:} Random ordering of
pseudo-labels for subquestions resulted in a large performance drop,
which sharply contrasts with the findings observed in few-shot language-only
scenarios \cite{min2022rethinking}. This divergence could stem from
the fact that subquestions are strongly connected, resulting in a
structured label space for each question. Shuffling the labels could
potentially disrupt this structural coherence. Another plausible explanation
could be the intrinsic contextual nature of Visual QA, wherein associating
answers with their corresponding subquestions provides more meaningful
local reasoning patterns.\footnote{We did not put the image before the subquestion-answer pair to reduce
the context length.}. Introducing randomness to the pairing might have disrupted the coherent
deliberative patterns. This effect is further emphasized by the considerably
more pronounced decline in performance when pairing subquestions with
placeholder answers. In such instances, LVLMs lack semantic reasoning
analogies to draw upon, thus contributing to the major drop in performance.

\subsubsection{Limitations}

We caution the readers that our study used a single open LVLM (for
visual-linguistic QA) and an open LLM (for question decomposition),
namely the recent OpenFlamingo and Vicuna 13B, respectively. However,
the $\ModelName$ framework is generic and can be applied with any
LVLMs. Since OpenFlamingo is based on Flamingo \cite{alayrac2022flamingo},
it inherits weaknesses of the original model including sporadic instances
of hallucinations and ungrounded assumptions. Moreover, it may struggle
to generalize effectively to sequences longer than those encountered
during training. The computational cost of inference does not scale
too well with the number of shots. We acknowledge that the outcomes
of our study are not yet comparable to the state-of-the-art supervised
learning accompanied by domain knowledge. Nevertheless, as LVLMs continue
to advance in sophistication in the foreseeable future, there is a
possibility that these limitations will be mitigated.

\subsubsection{Conclusion}

We have introduced $\ModelName$, a novel multimodal prompting framework
designed to guide in-context few-shot learning in pretrained Large
Vision Language Models (LVLMs), focusing on complex compositional
Visual QA tasks. This approach holds the promise of a novel training-free,
inference-only strategy that facilitates rapid deployment. Through
the utilization of $\ModelName$, we demonstrated that query-focused
semantic sampling of demonstrations plays a crucial role in Visual
QA, highlighting the difference between language-only and vision-language
settings. To devise deliberative reasoning demonstrations, we decompose
a compositional question into subquestions, each of which is automatically
labelled using only the label for the original question. We demonstrate
the efficacy of $\ModelName$ coupled with OpenFlamingo on large-scale
Visual QA datasets, namely GQA, GQA-OOD, CLEVR, and CRIC. The results
highlight the critical roles of sampling in the neighborhood of the
image, the decomposition of complex questions, and the precise pairing
of the subquestions and labels. These findings offer new insights
in vision-language settings.

{\small{}\bibliographystyle{ieee_fullname}
\bibliography{longdang.bib,ME.bib,truyen.bib}
}{\small\par}

\section{Biography Section}
\begin{IEEEbiography}[{\textbf{{\includegraphics[width=1in,height=1.25in]{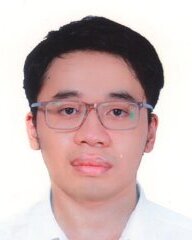}}}}]{Long Hoang Dang}
\textbf{ }received the PhD degree in Applied Artificial Intelligence
Institute, Deakin University, Australia, in 2024. He is currently
a Lecturer at the Faculty of Information Technology, Posts and Telecommunications
Institute of Technology (PTIT), Hanoi, Vietnam. His research interests
include image and video processing, machine learning and computer
vision, especially vision and language reasoning.
\end{IEEEbiography}

\textbf{\vspace{3pt}
}
\begin{IEEEbiography}[{\textbf{{\includegraphics[width=1in,height=1.25in]{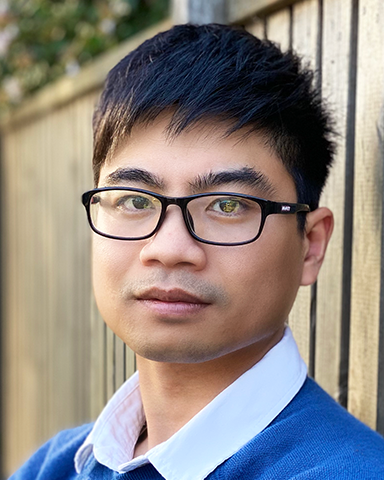}}}}]{Thao Minh Le}
\textbf{} is a Research Fellow at the Applied Artificial Intelligence
Institute, Deakin University. His research interests focus on deep
learning for computer vision tasks, and exploring the interplay between
vision and language. His other research topics include machine learning
for health and large-scale data analytics across different input modalities. 
\end{IEEEbiography}

\textbf{\vspace{3pt}
}
\begin{IEEEbiography}[{\textbf{{\includegraphics[width=1in,height=1.25in]{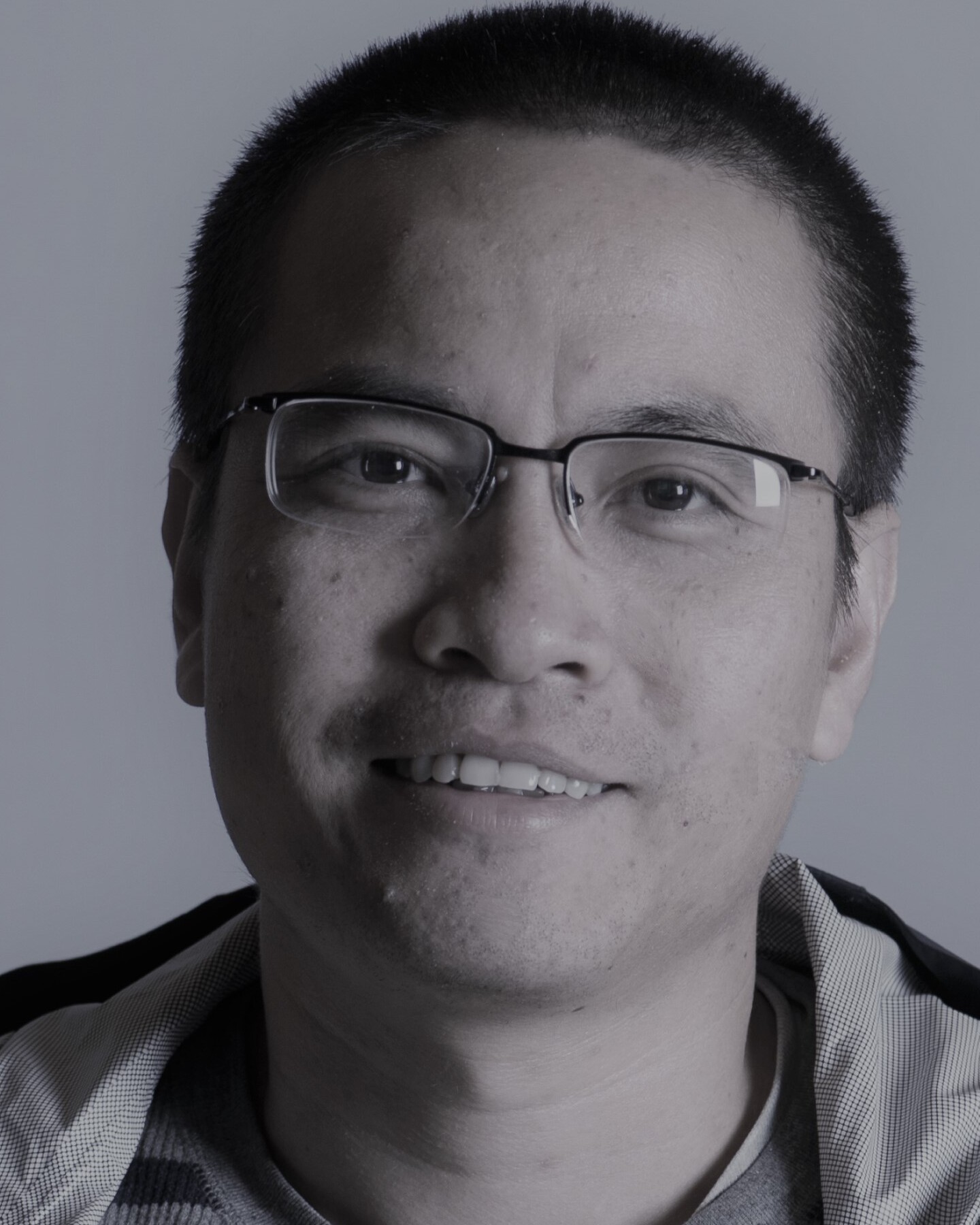}}}}]{Vuong Le}
\textbf{} is a Machine Learning Scientist at Amazon Australia. He
is also holding an affiliation with the Applied Artificial Intelligence
Institute at Deakin University. His research interests include multimodal
reasoning, especially on video understanding and human behavior analysis.
Dr. Le\textquoteright s works have resulted in one monograph, one
book chapter and more than thirty scholarly articles published at
major machine learning and computer vision journals and conferences,
including the top venues of the field: CVPR, ICCV, ECCV, IJCV, IJCAI.
His works have been cited more than 3200 times making up to an h-index
of 19. He is also holding eleven U.S. patents on related topics.
\end{IEEEbiography}

\textbf{\vspace{3pt}
}
\begin{IEEEbiography}[{\textbf{{\includegraphics[width=1in,height=1.25in]{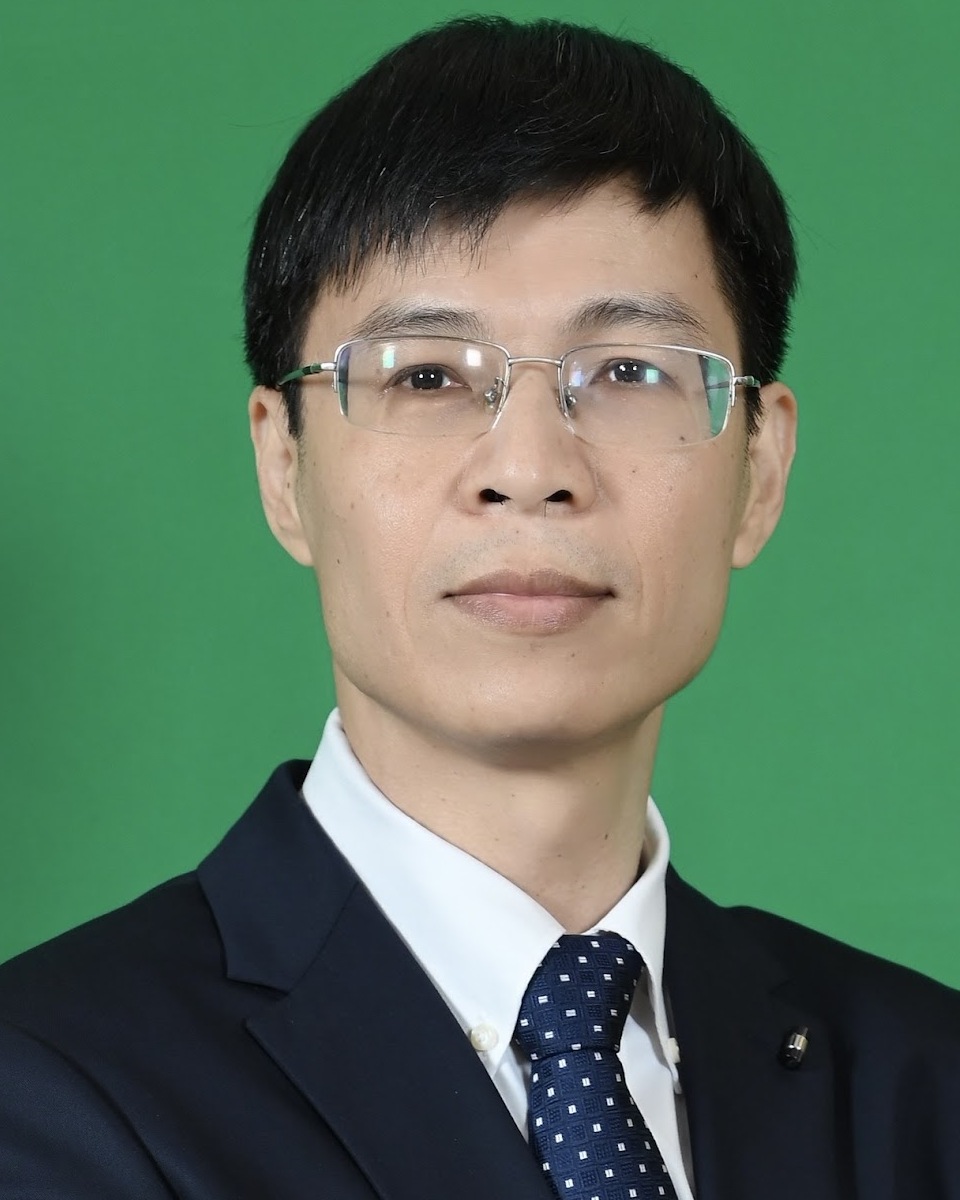}}}}]{Tu Minh Phuong}
\textbf{ }received the PhD degree in Control in technical systems
from National Academy of Sciences, Uzbekistan, in 1995. He joined
Faculty of Information Technology, Posts and Telecommunications Institute
of Technology (PTIT), Hanoi, Vietnam, in 2000, and became Professor
of computer science in 2019. He is Chairman of University Council
at PTIT. His research interests include machine learning, recommender
systems, natural language processing, and bioinformatics.
\end{IEEEbiography}

\textbf{\vspace{3pt}
}
\begin{IEEEbiography}[{\textbf{{\includegraphics[clip,width=1in,height=1.25in]{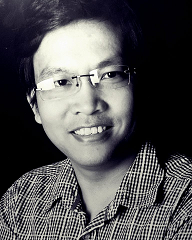}}}}]{Truyen Tran}
\textbf{ }is Associate Professor at Deakin University where he is
pushing the AI boundaries of deep learning, machine reasoning, unifying
language and vision, cognitive architectures and artificial social
intelligence. With his role as the Head of AI, Health and Science,
he is also driving the effort for transforming science, healthcare
and engineering through AI. Dr Tran has received multiple international
recognition, awards and prizes for his research contributions. He
holds a BSc. from the University of Melbourne (2001) and a PhD in
Computer Science from Curtin University (2008).
\end{IEEEbiography}

\end{document}